\documentclass[prodmode,acmtmis]{acmsmall} 

\usepackage[ruled]{algorithm2e}
\usepackage{caption}
\usepackage{amssymb}
\usepackage{epstopdf}
\usepackage{multirow}
\usepackage{color}
\usepackage{fancyhdr}
\usepackage{amsmath}
\usepackage{enumerate}
\usepackage{placeins}
\usepackage{color}

\newcommand{\CS}[0]{\mathit{CS}}
\newcommand{\TWS}[0]{\mathit{TWS}}
\newcommand{\phico}[0]{\mathit{\phi-Coefficient}}
\newcommand{\PS}[0]{\mathit{PS}}
\newcommand{\OR}[0]{\mathit{OR}}
\newcommand{\YuleQ}[0]{\mathit{YuleQ}}
\newcommand{\IG}[0]{\mathit{IG}}

\renewcommand{\algorithmcfname}{ALGORITHM}
\SetAlFnt{\small}
\SetAlCapFnt{\small}
\SetAlCapNameFnt{\small}
\SetAlCapHSkip{0pt}
\IncMargin{-\parindent}

\begin{document}


\title{Business Process Deviance Mining: Review and Evaluation}

\author{HOANG NGUYEN
	\affil{Queensland University of Technology, Australia}
	MARLON DUMAS
	\affil{University of Tartu, Estonia}
	MARCELLO LA ROSA
	\affil{Queensland University of Technology, Australia}
	FABRIZIO MARIA MAGGI
	\affil{University of Tartu, Estonia}
	SURIADI SURIADI
	\affil{Queensland University of Technology, Australia}}

\begin{abstract}
Business process deviance refers to the phenomenon whereby a subset of the executions of a business process deviate, in a negative or positive way, with respect to its expected or desirable outcomes. 
Deviant executions of a business process include those that violate compliance rules, or executions that undershoot or exceed performance targets. 
Deviance mining is concerned with uncovering the reasons for deviant executions by analyzing business process event logs.
This article provides a systematic review and comparative evaluation of deviance mining approaches based on a family of data mining techniques known as sequence classification.
Using real-life logs from multiple domains, we evaluate a range of feature types and classification methods in terms of their ability to accurately discriminate between normal and deviant executions of a process. 
We also analyze the interestingness of the rule sets extracted using different methods. 
We observe that feature sets extracted using pattern mining techniques only slightly outperform simpler feature sets based on counts of individual activity occurrences in a trace.
\end{abstract}

\terms{Process Mining, Business Process Deviance, Sequence Classification}



%
%

\maketitle

\section{Introduction} 
Process mining is a family of techniques to extract knowledge of business processes from event logs~\cite{ProcessMiningBook}. It encompasses, among others, techniques for automated discovery of process models from event logs, techniques for checking conformance between a given process model and an event log, as well as techniques for analyzing and predicting performance of business processes based on event logs.


This paper deals with \emph{business process deviance mining} -- a family of process mining techniques aimed at analyzing event logs in order to explain the reasons why a business process deviates from its normal or expected execution. Such deviations may be of a negative or of a positive nature -- cf.\ theory of positive deviance~\cite{Gretchen2004}. Positive deviance corresponds to executions that lead to high process performance, such as achieving positive outcomes with low execution times, low resource usage or low costs. Negative deviance refers to the executions of the process with low process performance or with negative outcomes or compliance violations.

The input of deviance mining is an event log consisting of a set of labelled traces (the so-called \emph{training set}). Each trace represents the execution of one case of the business process under analysis. It consists of a sequence of events, where an event corresponds to the execution of an activity. Each trace is associated with a label indicating whether the trace is ``normal'' or ``deviant''.

Given this input, the problem of deviance mining is that of calculating a function (called a \emph{classifier}) that takes as input a trace and outputs a class for this trace (normal or deviant). Such function must produce accurate labels, i.e.\ it should guess the correct class of a trace both for traces in the training set but also for other unseen traces. Furthermore, since the purpose of deviance mining techniques is to explain deviance, their output should be expressed in terms of patterns or rules that allow an analyst to extract useful insights regarding the sources of the observed deviance.

Since traces consist of sequences of events, one family of techniques applicable for deviance mining is \emph{sequence classification}~\cite{Xing2010Survey}. The goal of sequence classification is to construct classifiers that discriminate between two or more classes of sequences. One key issue in sequence classification is that of extracting features from the sequences that can be given as input to standard classification techniques such as decision trees. Such features can be extracted for example using sequence mining techniques. Various such techniques have been studied in the context of business process deviance mining as discussed later. However, no comparative study has been conducted to assert which techniques are more suitable in this setting. 

This article presents a comparative evaluation of sequence classification techniques for business process deviance mining. Based on a review of the state of the art, the article identifies two families of techniques that have been employed for deviance mining -- those based on frequent pattern mining and those based on discriminative pattern mining. Representatives of these two families are benchmarked using a battery of real-life event logs, covering situations where deviance is frequent (balanced datasets) and others where deviance is rare (unbalanced). These representative techniques are assessed in terms of \emph{accuracy} and \emph{interestingness} measures. The former measures serve to quantify the extent to which the extracted patterns correctly reflect the classification of cases into normal and deviance, whereas the latter measures serve as a proxy for the usefulness of the extracted patterns.

This article is an extended version of a conference paper~\cite{Nguyen2014}. With respect to the conference paper, this article adds: (i) a systematic classification and review of deviance mining techniques; (ii) a refinement of the evaluated techniques using Fisher score for feature selection; (iii) an extended evaluation with three additional event logs; and (iv) an assessment of the interestingness of the extracted classifiers in addition to their accuracy.

The paper is structured as follows. Section~\ref{sec:StateOfTheArtReview} discusses existing methods for business process deviance mining. Section~\ref{sec:EvaluationSetup} outlines the methods for feature extraction and for classification evaluated in this study. Next, Section~\ref{sec:AccuracyEvaluation} presents the experimental results in terms of classification accuracy while Section~\ref{sec:RulesInterestingnessEval} presents the results in terms of rules interestingness. Section~\ref{sec:Conclusion} summarizes the contribution and discusses directions for future work.  
\section{State of the Art}\label{sec:StateOfTheArtReview}
To analyze the state of the art in the field of sequence mining for business process deviance mining, we started by conducting a literature search using the systematic review principles of~\cite{Kitchenham2004}. The search process started by submitting queries to a well-known literature database, Google Scholar.\footnote{\url{http://scholar.google.com}} with keywords
associated with the scope of the paper. We constructed queries by combining the keyword ``business process'' with the following keywords: ``deviance mining'' (referring to the problem under study), ``sequence mining'' (the broad class of solutions of interest) and ``discriminative patterns'' (the specific class of solutions of interest). Since the term ``workflow'' is sometimes used as a quasi-synonym of ``business process'', we also included queries combining ``workflow'' with the above keywords.
For each query, we gathered the first 20 hits in Google Scholar.
As we conducted the search, we noted additional terms appearing in the titles
of relevant papers, namely  ``sequential patterns'' and ``signature patterns''
and also used such terms in the search.
The search was performed in September 2015. It resulted
in 200 hits. Based on title, we filtered out papers that were clearly out of scope and removed duplicates.
This filtering reduced the number of candidate papers to 34.
We then proceeded with an inspection of the references of each paper in order to find other papers related to the topic. This activity increased the number of candidate papers to 48.

\begin{figure}[htb!]
	\centerline{\includegraphics[width=\textwidth]{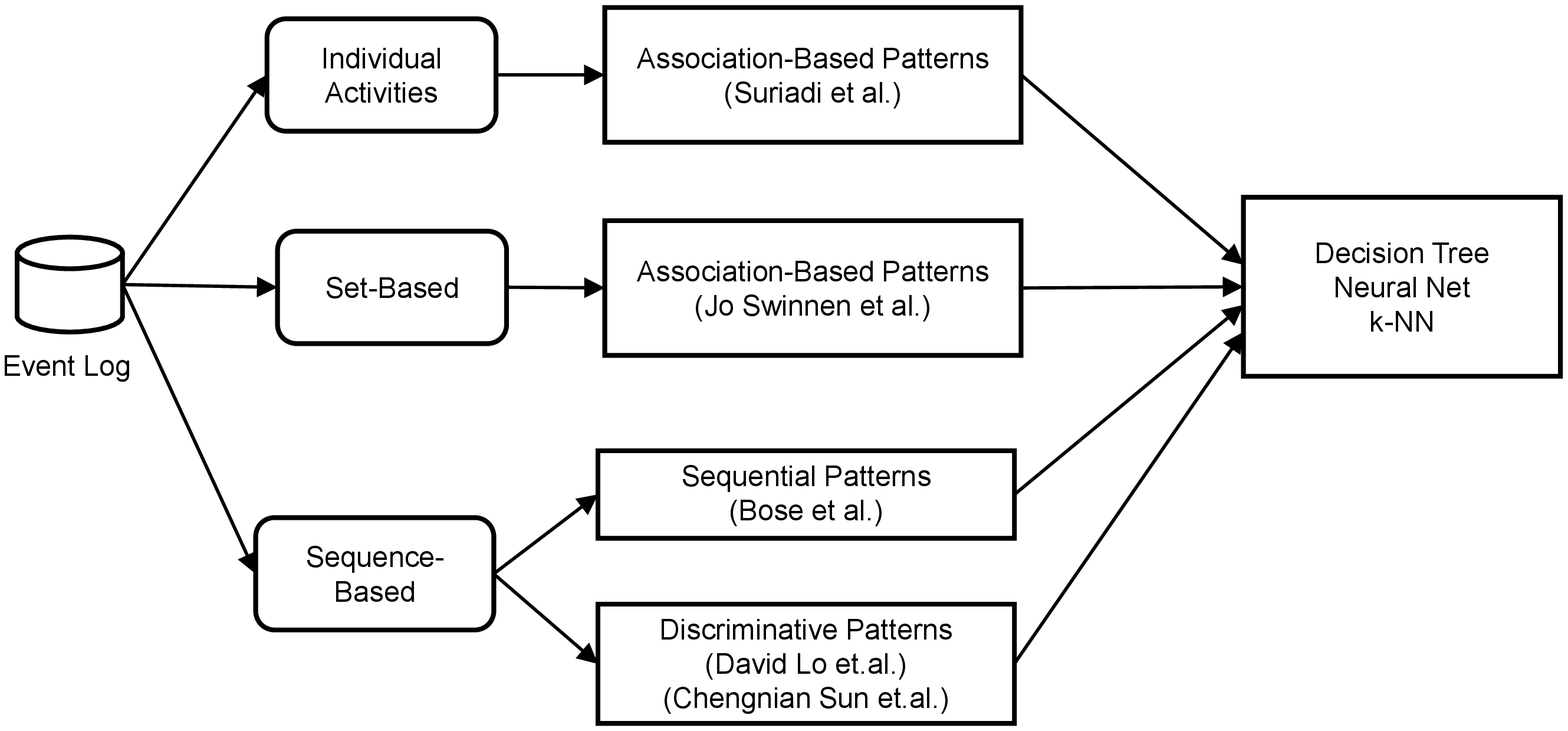}}
	\caption{Taxonomy of Business Process Deviance Mining Approaches}
	\label{fig:one}
\end{figure}

From these publications we could classify them into three main groups: (1) pattern mining papers which focus on discovery of certain interesting patterns from datasets, (2) classification papers which propose new techniques and algorithms to improve classification quality with regards to a target class variable, and (3) pattern-based classification papers which involve mining patterns and evaluating the predictive power of the patterns with regards to a target class variable. Note that papers in all groups span related work within and outside the domain of business processes. Our review was then concentrated on the last group of 9 papers since we wanted to survey related techniques which have been evaluated in or can be applied for business processes. From this group, we selected two representative papers of automated pattern mining and classification evaluation. The first one is~\cite{Lo2009} which examined discriminative patterns in the field of software engineering, and the second one is~\cite{bose13DiscoveringSignaturePatterns} which examined different types of sequential patterns in business processes. In addition, based on our knowledge, we also added two case study papers which proposed manual but closely related approaches:~\cite{Suriadi2013} examined business process activities and their effects on the process duration, and~\cite{Swinnen2012} examined the predictive power of sets of activities. Consequently, our survey could identify a taxonomy (see Fig.~\ref{fig:one}) represented by four papers and consisted of three categories of techniques: (1) based on individual activities, (2) set-based, and (3) sequence-based. In addition sequence-based techniques could be further distinguished into sequential and discriminative.

\FloatBarrier

All automated techniques for deviance mining (i.e.\ excluding manual delta analysis), involve a pattern extraction phase  where patterns are extracted from traces seen as sequences of simple symbols (tokens) representing activity occurrences, followed by a classifier construction phase, where each trace is abstracted as a vector of features, and these vectors are given as input to a classification method such as decision tree miner, which produces a \emph{classifier}. The resulting classifier is then analyzed in order to extract individual patterns or rules combining multiple patterns, that explain a high percentage of the observed deviance. These patterns or rules are given as input to analysts to help them understand the sources of observed deviance in the process.

Different techniques differ depending on the abstraction of the event logs used to define the features. The three ways of abstraction as shown on the taxonomy are described as follows.
\begin{enumerate}
	\item Occurrence count of \emph{individual activities} in a trace as in~\cite{Suriadi2013}. In this feature extraction method, each activity type appearing in the log (e.g.\ ``Receive Purchase Order Change'', ``Issue Invoice'', ``Invoice Paid'') is treated as a numerical feature. For a given trace, the value of an activity feature is the number of times the activity in question occurs in the trace. For example if in a given trace, activity ``Receive Purchase Order Change'' appears 3 times, this is the value of the corresponding activity feature. This approach is quite straightforward with consideration that single or multiple activities may have significant influence on the case outcome.
	
	\item Frequent \emph{set of trace attributes} as in \cite{Swinnen2012}. In this feature extraction method, frequent sets of activities are taken as a feature regardless of the order of their occurrence in the trace. A trace in this view follows the market basket data model in data mining. Frequent item set mining algorithms, such as Apriori \cite{agrawal1994fast}, can then be applied to mine patterns that meet a support and/or confidence threshold. This approach takes interest in the association among activities \emph{regardless of their ordering}.
	
	\item \emph{Sequence of events} that occurs multiple times in a trace or across traces. These features represent different way of event coordination such as loop, sub-processes or synchronization. We select two representative types of sequence-based features: one from \cite{Bose2009} (called Sequential Patterns) and one from \cite{Lo2009} (called Discriminative Patterns).
	\begin{enumerate}
  \item \cite{Bose2009} proposes sequential patterns including \emph{Tandem Repeats}, \emph{Alphabet Tandem Repeats}, \emph{Maximal Repeats} and \emph{Alphabet Maximal Repeats}. These patterns capture different types of control-flow relations in a business process (loops, parallelism and subprocesses) and have been shown to be potentially suitable for analyzing business process deviance ~\cite{bose13DiscoveringSignaturePatterns}. More specifically, tandem repeats mining searches for recurrent sequences of events within a trace which represent loops, whereas maximal repeats mining searches for any repetitive sequences in the whole log which represent sub-processes. An alphabet-type feature represents multiple tandem repeats or maximal repeats, respectively, if they share the same constituent activities (alphabet is the set of unique activities constituting a sequential pattern). Alphabet-based patterns, thus, capture variations in tandem repeats or maximal repeats due to parallelism.

  \item \cite{Lo2009} proposes iterative features consisting of consecutive or non-consecutive events that have multiple occurrence within a trace or across traces. This technique is different to tandem/maximal repeats in the sense that the atomic events in an iterative feature do not necessarily occur close together, which captures synchronized behaviors among events. In addition, \cite{Lo2009} uses feature selection technique to return only features that are strongly associated with the outcome classes (so called "Discriminative Patterns"). This association is measured by a weight value, such as \textit{Fisher score} \cite{duda2012pattern}.
\end{enumerate}
\end{enumerate}

The first method (\emph{individual activities}) is in essence a baseline. Indeed, the occurrence of a single activity in a trace is the simplest form of feature one can extract from a sequence.
In this work, we aim at evaluating how much added-value other feature extraction methods (i.e.\ set-based and sequence-based variations) could contribute on top of the \emph{individual activities}. Thus, we study seven feature sets in the sequel: (i) individual activities (IA); (ii) maximal repeats (MR); (iii) alphabet maximal repeats (MRA); (iv) tandem repeats (TR); (v) alphabet tandem repeats (TRA); (vi) iterative patterns (IP); and (vii) set-based patterns (SET).

\smallskip
Deviance mining is partially related to predictive monitoring of business processes~\cite{PredictiveMonitoring,MetzgerLISFCDP15}. Whereas deviance mining aims at explaining the reasons for deviant cases in \emph{post mortem} manner, predictive monitoring aims at predicting whether or not an ongoing case will end up in a normal or a deviant category upon its completion -- for example predicting whether or not an ongoing case will lead to a customer complaint or a deadline violation. Deviance mining and predictive monitoring techniques have overlapping inputs but distinct outputs. Deviance mining techniques take as input a set of traces classified into normal and deviant, and returns a set of rules that explain the deviance. Predictive process monitoring techniques additionally take as input the incomplete trace  of a running case of the process, and they return a prediction of the outcome of this case.

Given their overlapping inputs, predictive monitoring and deviance mining techniques have common concerns. Specifically, they both have to deal with the question of how to extract features from the set of historical traces in order to construct a classifier -- with the difference that in predictive monitoring, one has to take into account the fact that the classification is done with respect to the trace prefix of an incomplete case, and thus the ``state'' of execution of the current case needs to be taken into account. Not surprisingly, some of the feature extraction methods mentioned above can also be found in the context of predictive monitoring. For example, \cite{PredictiveMonitoring} considers both individual activity features as well as frequent sequential patterns, whereas \cite{LeontjevaCFDM15} extends these features with others, including the index in the trace where a given activity occurs (e.g. does a given activity occur in the first or second position in the trace?) as well as other features extracted from hidden markov models constructed from the historical traces. \cite{LeoniAD16} provides a framework for feature extraction that additionally considers the counts of individual activity occurrences and their index as possible features, as well as a range of other features related to the data-flow and the resource perspective (e.g.\ current value of a given variable of the process, and current workload of a given resource). Similar features are also employed in \cite{ConfortiLRA13,ConfortiLRAH15} to predict process risks. With respect to these latter two works, the present article focuses on studying the predictive power of control-flow features. On the other hand, it goes deeper by considering not only individual activity features but also set-based and sequence-based patterns as discussed above.
\section{Evaluation Setup}\label{sec:EvaluationSetup}
This section presents the evaluation setup. First, it describes the datasets used; next it outlines the methods used for feature extraction and for classification.

\subsection{Datasets}

We used six datasets derived from five real-life event logs. We selected these datasets in order to cover classifications of ``normal'' and ``deviant'' cases based on \emph{temporal} and \emph{non-temporal} criteria. Note that \emph{case} and \emph{trace} are used interchangeably in this paper, as well as \emph{feature} and \emph{pattern}. In the former criterion, the difference between deviant and normal cases is made on the basis of the process duration w.r.t. a duration threshold (i.e. slow vs. normal processes); in the latter criterion, the difference between types of cases depend on an outcome attribute of the case (e.g.\ failing or successful outcome).

\begin{table}[t]	
{\footnotesize{
\begin{center}
	\begin{tabular}{|l|c|c|}
		\hline \textbf{Dataset}	& \textbf{Deviance criterion}	& \textbf{Deviant class distribution}\\
		\hline Schedule			&	non-temporal 				& 50\%	\\
		\hline MySQL			&	non-temporal 				& 50\%	\\ 
		\hline Hospital1		&	temporal 					& 45\%	\\ 	
		\hline Hospital2		&	temporal 					& 44\%	\\ 
		\hline Insurance1		& 	temporal 					& 62\%	\\ 
		\hline Insurance2		& 	non-temporal 				& 71\%	\\		
		\hline
	\end{tabular}
\end{center}
}}
  \caption{Labeling of the six datasets used for the evaluation.}\label{tab:datasets1}
\end{table}


\begin{table}[t]
{\footnotesize{
\begin{center}

\begin{tabular}{|l|c|c|c|c|c|c|c|}
\hline 
\multirow{3}{*}
{\textbf{Dataset}} 	& \textbf{Normal}	& \textbf{Deviant}	& \textbf{Total}	& \textbf{Total}	& \textbf{Mean}				& \textbf{Mean}	\\
 					& \textbf{cases}	& \textbf{cases}	& \textbf{cases}	& \textbf{event}	& \textbf{event classes}	& \textbf{events}	\\
  					& \textbf{}  		& \textbf{} 		& \textbf{} 		& \textbf{classes} 	& \textbf{per case}			& \textbf{per case} \\

\hline Schedule		& 2,140	& 2,140	& 4,280	& 19 	& 15	& 124	\\
\hline MySQL 		& 51	& 51	& 102	& 16	& 16	& 24	\\
\hline Hospital1	& 448	& 363 	& 811	& 26	& 14	& 18	\\
\hline Hospital2 	& 15,929	& 12,847	& 28,776	& 32	& 8		& 10	\\
\hline Insurance1	& 1,921 & 3,195	& 5,116	& 13	& 7		& 20	\\
\hline Insurance2	& 79 	& 197	& 276	& 19	& 6		& 16	\\
\hline

\end{tabular}
\end{center}
}}
  \caption{Descriptive statistics for the six datasets.}\label{tab:datasets2}
\end{table}
The six logs fall into two sets. The first set (including the first two logs in Table~\ref{tab:datasets1} and~\ref{tab:datasets2}) is recorded software behaviors \cite{Lo2009}. These two logs contain traces in which every event is a software command. Every trace can be viewed as a series of commands executed while using the software. The trace outcome is marked as ``successful'' when the execution has no errors, and ``failing'' when the execution results in an error (deviant case). The Schedule log was obtained from a Siemens system and the MySQL log was obtained from the a MySQL database engine. These processes are machine-based and highly repeatable.

By contrast, the second set (including the four remaining logs in Tables~\ref{tab:datasets1} and~\ref{tab:datasets2}), contains weakly structured processes with intensive human involvement. The Hospital1 log pertains to a chest pain patient flow in an emergency department of an Australian hospital. Traces are labelled as ``quick'' for cases that complete within a 180 minutes, and``slow'' (or deviant) otherwise. 

The Hospital2 log is provided by another Australian hospital pertaining to the assessment process of mental health patients from the time the patients are referred to the hospital (as outpatients) to the time they are first assessed by qualified medical personnel. A case is labelled as ``quick'' if the assessment process is completed within 5 days, or ``slow'' (or deviant) otherwise. 

The Insurance1 log comes from an Australian insurance company and records an extract of the instances of a commercial insurance claims handling process. Also for this log, we used a temporal deviance criterion for classification, with ``quick'' being the label for those cases that complete within 30 days, and ``slow'' otherwise. Also in this case, slow cases are deviant cases.

The Insurance2 log is extracted from the Insurance1 dataset to evaluate the relationship between individual activities and non-temporal outcome. The two outcome labels, ``mistake'' and ``rejected'', are non-temporal that reflect the final decision on cases (the latter is deviant). The two activities ``Authorise Decline'' and ``Confirm Decline'' are also deliberately filtered out.

\subsection{Model construction}

The model construction has two stages: pattern mining and model training/testing. Given a labeled event log (i.e. a log where each trace is labeled as ``normal'' or ``deviant''), the first stage will mine and select features. The latter stage will then construct a classifier by transforming the training traces into a vector space and using standard classification techniques. In this context, a trace $t$ is converted to a vector of features $(\langle\: f_1, ... f_n \:\rangle, l)$, where $f_i (i=1,2,...,n)$ is the value of the $i^{th}$ feature for trace $t$ and $l$ is the label (normal or deviant). The value is the frequency count of the feature in the trace.

We use a feature selection technique as used in \cite{ChengDPMine2008} and \cite{Lo2009}. This technique selects a feature based on a weight of its statistical correlation with the outcome classes. The weight can be, for example, \textit{information gain} or \textit{Fisher score}. Fisher score is used in this paper in the same way as \cite{Lo2009}. Basically, in the first place, this technique selects an initial set of frequent features based on a support (frequency) level. Then, Fisher scores are computed for all features. Finally, it selects a final set of features by sequentially checking each feature from a high to low Fisher score ranking and selecting one that covers at least one trace from the trace database.

In this technique, the number of features selected is controlled by two parameters: \emph{minimum support level} and \emph{coverage threshold}. According to \cite{ChengDPMine2008}, the support level has an impact on the discriminative power. A minimum support level of 0.25 is used for all feature types. The coverage threshold $\theta$ is used in the final selection step, in the way that any trace covered by more than $\theta$ features will be excluded from being checked against the next features for coverage. The aim is the final feature set will contain high-score features that can cover as many traces as possible, without being overly concentrated on a small proportion of traces. A common coverage threshold of 5 is used for all feature types.

In the second stage, to construct classifiers from the labeled samples, we can use a range of methods. Given that we seek an explainable classifier, a natural choice is decision tree learning, which produces trees from which human-readable rules can be extracted. This is the approach employed for example in~\cite{Suriadi2013,bose13DiscoveringSignaturePatterns,Sun2013}. Hence, we include decision trees (C4.5 method implemented in RapidMiner) as a baseline in the evaluation. We also include a k-NN (k-Nearest Neighbors) classification method in the evaluation as an example of a simple classification technique that does not construct an explicit classification model, but instead classifies a given sample based on its most similar samples. Specifically, given a sample, the $k$ nearest neighbors are found and the class (normal or deviant) that is most common among neighbors in the training set is used to classify the given sample. We set the value of parameter $k$ to 8, after initial trial-and-error to find a value yielding higher accuracy. Although no rules can be extracted from a k-NN classifier, the output of k-NN is explainable in the sense that given a trace $t$, one can show which ``similar'' traces have been used to classify trace $t$ as normal or deviant. Finally, we included neural networks in the evaluation as representative of a method that can adjust itself to the data and handle large feature sets~\cite{NeuralNetworks} even though it does not produce explicit (understandable) rules as decision trees do.

Different tools are used for feature mining and classification. Tandem/maximal repeats and their alphabet variants are mined using the Signature Discovery plug-in in ProM \cite{bose13DiscoveringSignaturePatterns}, while iterative pattern mining is based on the implementation in \cite{Lo2009}. Set-based pattern mining is implemented using the FP-Growth algorithm for frequent item set mining \cite{han2000fpgrowth}. Tools for classification model training and testing are built on libraries and workflows of RapidMiner v6.0. 

Five-fold validation is used consistently from pattern mining to model training and testing. A log is randomly divided into five training and five testing datasets with 80/20 proportion of original traces, respectively. The class distribution in the training and testing data remains the same as that in the original data. The training dataset is \emph{only} used for mining patterns and producing the classification model, and the testing dataset is \emph{only} used for testing the trained model. This is, indeed, aligned with pattern-based comparative methods used in \cite{ChengDPMine2008}, \cite{Lo2009}, \cite{Deshpande2005frequent} and \cite{lee2011mining}.

The deviant class is usually rare in many datasets as compared to the non-deviant class. This skewed class distribution is unexpected because we want to discover deviant features and test their discriminative power. However, support-based frequent pattern mining tend to return only features of the dominant class; as a consequence, the training model built would be biased towards the dominant class. In this case, the prediction on the skewed data could be highly accurate but of little use in practice. To address this challenge, oversampling is used for training data to increase the population of the deviant class by randomly duplicating its traces. In this way, the training data of Schedule, MySQL and Hospital2 dataset have been balanced. 

\section{Accuracy Evaluation}\label{sec:AccuracyEvaluation}
We measured classification accuracy in terms of the standard notion of \emph{accuracy} defined as $\frac{tp + tn}{tp + tn + fp + fn}$, where $tp$ is the number of traces correctly classified as deviant (true positives), $tn$ is the number of traces correctly classified as normal (true negatives), $fp$ is the number of false positives and $fn$ is the No. of false negatives. Additionally, we also report on the \emph{Area Under the ROC Curve (AUC)} of each classifier. The AUC corresponds to the probability that a random negative sample is ranked higher than a random positive sample in the list of samples ranked from most likely to least likely to belong to the deviant class. AUC has been shown to be suitable for evaluating the accuracy of classification techniques both for balanced and unbalanced datasets~\cite{HeG09}. This choice makes the results more comparable across the datasets.

\begin{figure}
	\centering{\includegraphics[width=\textwidth]{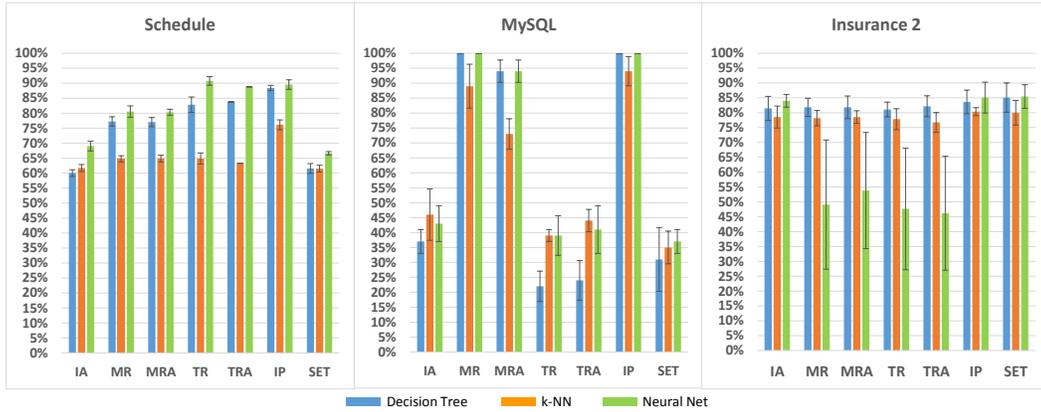}}
	\caption{Predictive Accuracy(1).}
	\label{fig:accuracy1}
\end{figure}

\begin{figure}
	\centering{\includegraphics[width=\textwidth]{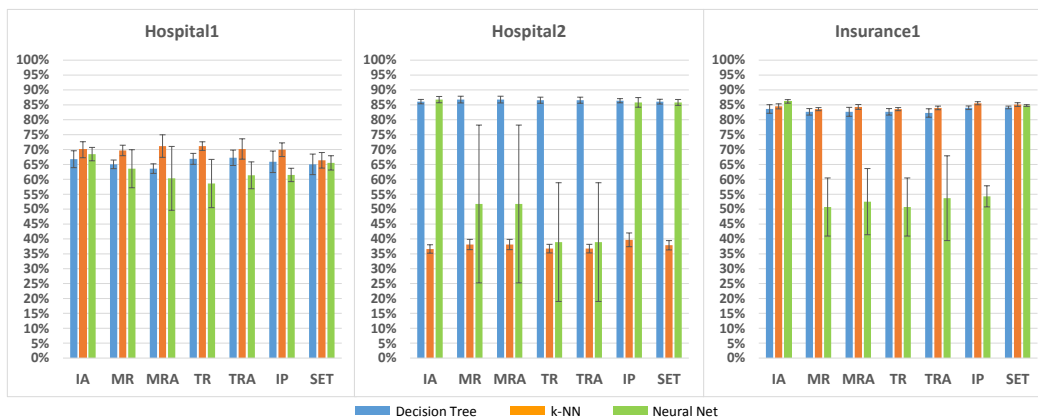}}
	\caption{Predictive Accuracy(2).}
	\label{fig:accuracy2}
\end{figure}

\begin{figure}
	\centering{\includegraphics[width=\textwidth]{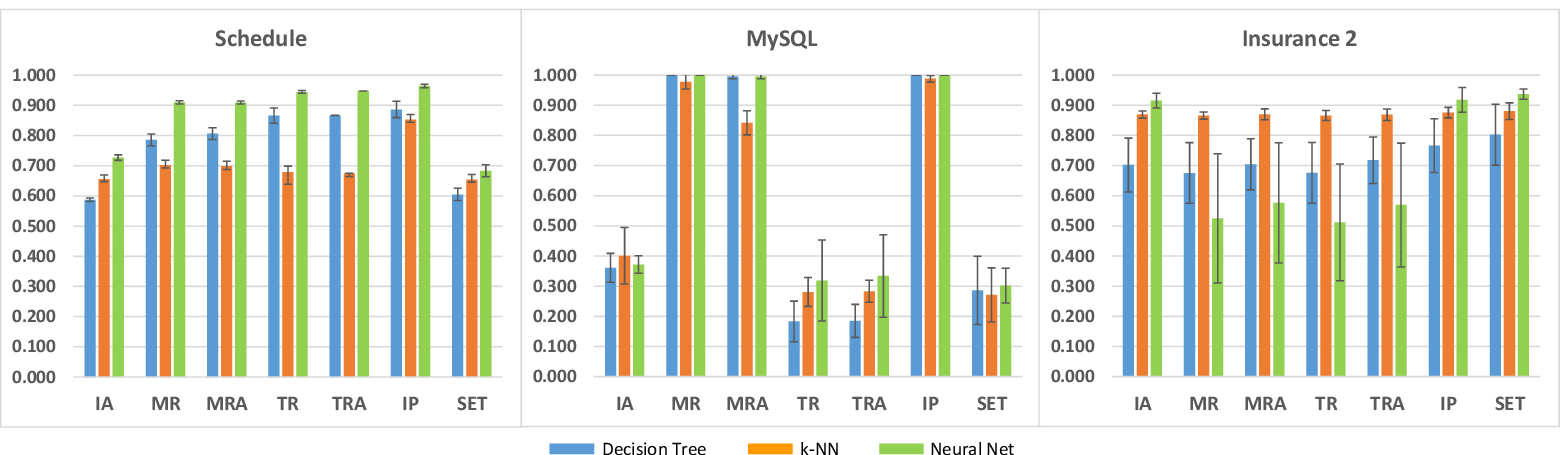}}
	\caption{AUC(1).}
	\label{fig:auc1}
\end{figure}

\begin{figure}
	\centering{\includegraphics[width=\textwidth]{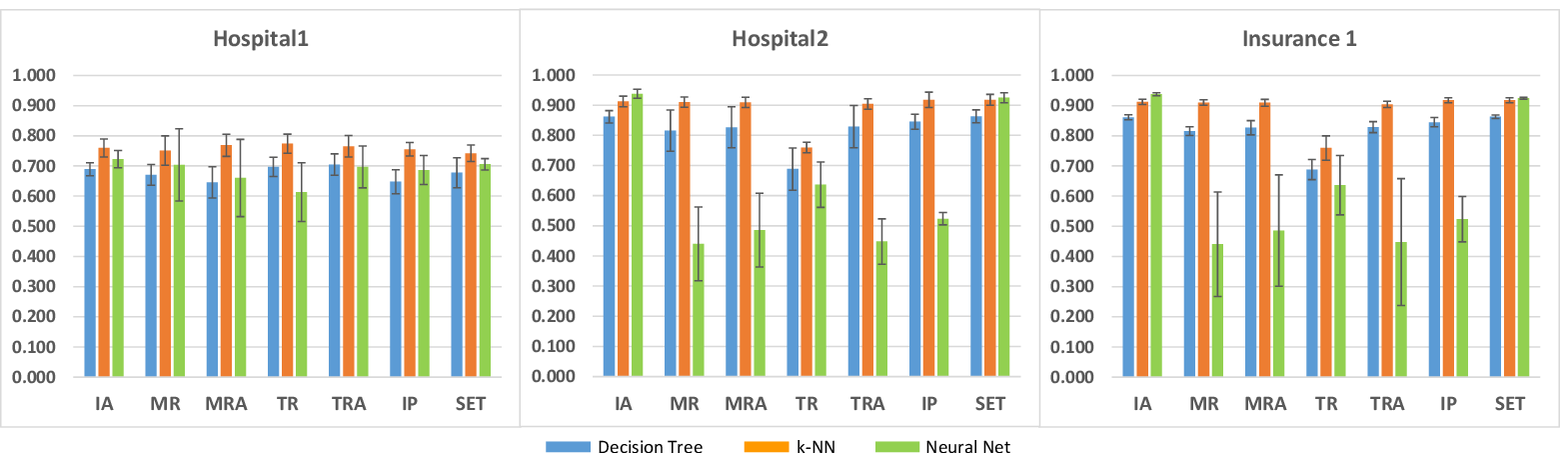}}
	\caption{AUC(2).}
	\label{fig:auc2}
\end{figure}

The prediction result is shown in Figure \ref{fig:accuracy1} - \ref{fig:auc2}. For the Schedule dataset, the IP, TR and TRA patterns provide the highest result (from 88 to 90\%). For the MySQL dataset, the MR and IP patterns score the best (100\%). For the Insurance2 dataset, the IP and SET patterns provide a slight improvement over the IA baseline (85.45\% vs. 84\% and 0.937 vs. 0.916). However, for other datasets (Hospital1, Hospital2 and Insurance1), we observed that the composite patterns have virtually no improvement over the IA.

\begin{table}[t]
	{\footnotesize{
	\begin{center}
	\begin{tabular}{|l|l|c|c|c|c|c|c|c|} \hline 
		
\multicolumn{2}{|c|}{\textbf{Dataset}} & 
\textbf{IA}	& \textbf{MR} 	& \textbf{MRA} 	& \textbf{TR} 	& \textbf{TRA} 	& \textbf{IP} 	& \textbf{SET} \\ \cline{1-9}

\multirow{2}{*} {Schedule}  
& Fisher Score & 
0.0381		& 0.056			& 0.061			& 0.022			& 0.026			& \textbf{0.376}			& 0.107 \\	\cline{2-9}
& No. of Features & 
17			& 45			& 39			& 46			& 28			& 31			& 93 	\\ \hline

\multirow{2}{*} {MySQL}  
& Fisher Score & 
0.002		& \textbf{$\infty$}			& 0.616			& 0.0016			& 0.0018			& \textbf{152,985}			& 0.003 \\	
\cline{2-9}
& No. of Features & 
5			& 9				& 7			& 4			& 3			& 13			& 9 	\\	\hline

\multirow{2}{*} {Hospital1}  
& Fisher Score & 
0.03		& 0.026			& 0.025			& 0.053			& 0.055			& 0.024			& 0.032 \\	\cline{2-9}
& No. of Features & 
21			& 26			& 28			& 11			& 11			& 66			& 100 	\\ \hline

\multirow{2}{*} {Hospital2}  
& Fisher Score & 
0.019		& 0.053			& 0.053			& 0.03			& 0.03			& 0.065			& 0.067 \\	\cline{2-9}
& No. of Features & 
32			& 9			& 9			& 2			& 2			& 21			& 22 	\\ \hline

\multirow{2}{*} {Insurance1}  
& Fisher Score & 
0.052		& 0.049			& 0.039			& 0.049			& 0.044			& 0.06			& 0.039 \\	
\cline{2-9}
& No. of Features & 
13			& 27			& 33			& 27			& 29			& 76			& 80 	\\ \hline 
\multirow{2}{*} {Insurance2}  
& Fisher Score & 
0.044		& 0.065			& 0.075			& 0.065			& 0.066			& 0.1			& \textbf{0.164} \\ \cline{2-9}
& No. of Features & 
18			& 21			& 22			& 19			& 18			& 31			& 35 	\\ \hline

	\end{tabular}
	\end{center}
	}}
	\caption{Average Fisher scores by feature types.}\label{tab:fisherscore}
\end{table}

Table \ref{tab:fisherscore} shows the average Fisher scores computed on the list of selected features for each feature type. Clearly, the features with better predictive power has higher average Fisher score than others. This is evident in the case of IP/Schedule and MR \&IP /MySQL (X/Y refers to a pair where X is the feature type and Y is the dataset). Likewise, IP/Insurance2 is highlighted with higher average Fisher score than the others.

\begin{figure}
	\centering{\includegraphics[width=\textwidth]{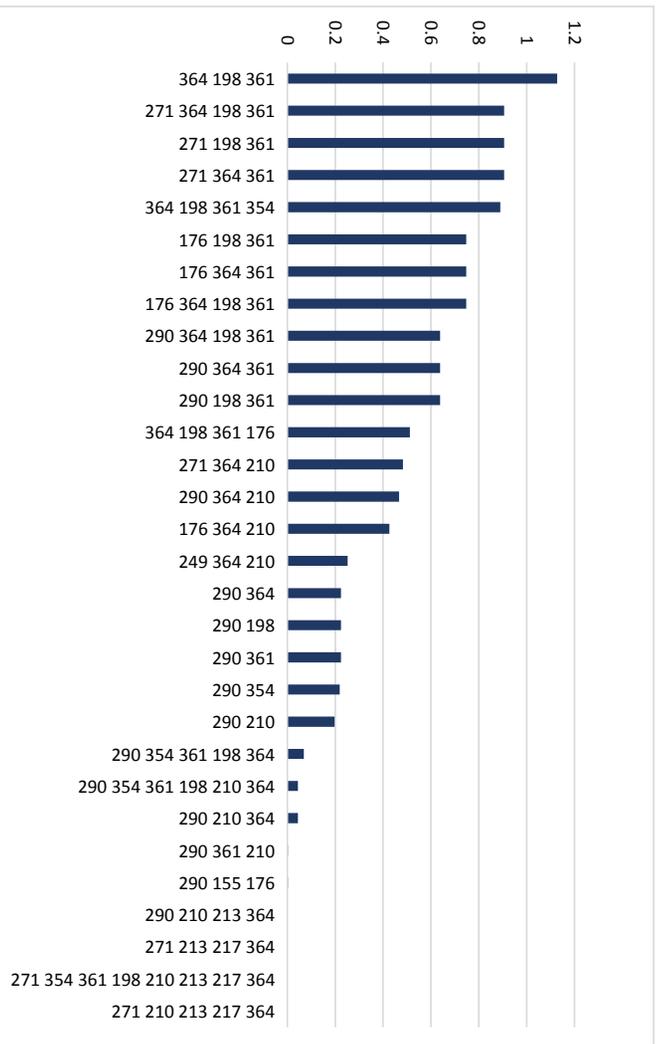}}
	\caption{Selected Features and Fisher scores (IP/Schedule). Every number represents one activity, e.g. `364' is one activity.}
	\label{fig:fisherscore1}
\end{figure}

\begin{figure}
	\centering{\includegraphics[width=\textwidth]{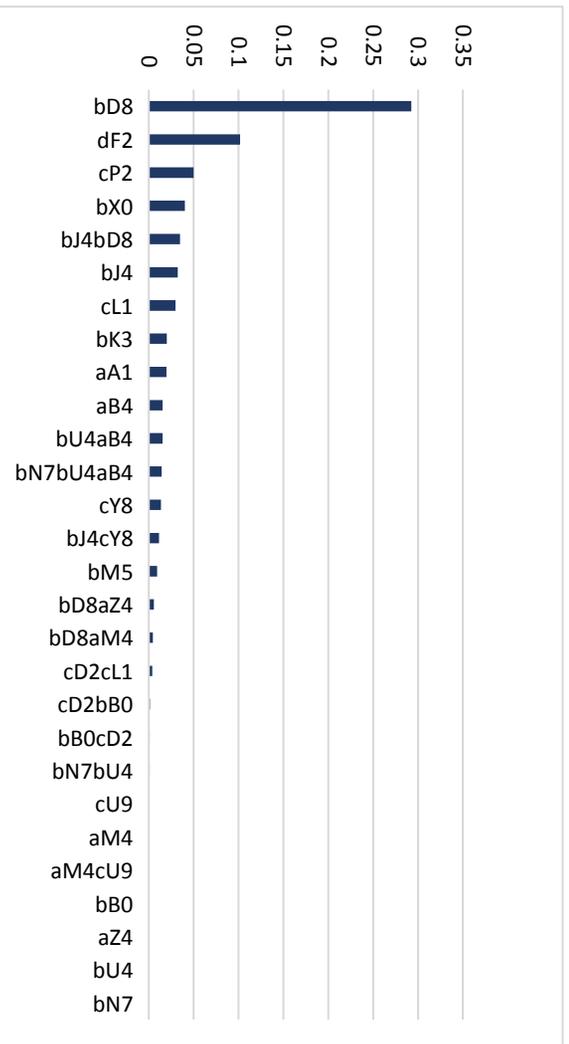}}
	\caption{Selected Features and Fisher scores (MR/Hospital1). The top-ranking encoded features such as bD8, cF2, cP2, and bx0 represent individual activities.}
	\label{fig:fisherscore2}
\end{figure}

The list of selected features ranked by Fisher scores indeed supports that feature types with higher predictive accuracy do provide a number of highest scored features. 
For example, IP/Schedule has quite many patterns with high scores (see Figure~\ref{fig:fisherscore1}) whereas IA/Hospital1 dominates the ranking (see Figure~\ref{fig:fisherscore2}\&\ref{fig:fisherscore2}).

\begin{figure}
	\centering{\includegraphics[width=\textwidth]{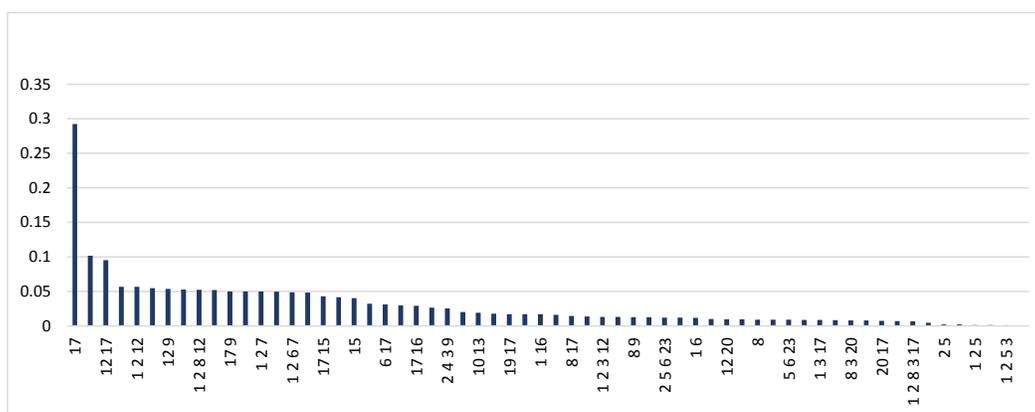}}
	\caption{Selected Features and Fisher scores (IP/Hospital1). Every numeric code represents one activity, e.g. `17' is one activity.}
	\label{fig:fisherscore3}
\end{figure}

Another observation is the composite patterns, i.e. non-IA, achieve higher accuracy than IA only when the case outcome is not temporal, e.g. Schedule and MySQL datasets. In other words, if the outcome is temporal, IA patterns likely carry most of the outcome-related signals since case duration is usually susceptible to the activity execution, i.e. the longer a case is, the more activities it has as observed in Hospital1, Hospital2 and Insurance1. Similarly, in the case of Insurance2, we deliberately use non-temporal outcome criterion and filter our related activities. As a result, the highest accurate feature type is SET.

In conclusion, three observations arise from our experiments. First, IA captures most of the signals if the deviance outcome is temporal. Second, composite patterns are highly predictive in case of structured processes, e.g. Schedule and MySQL datasets. Finally, in case of less structured processes, composite patterns yield negligible improvement over IA, possibly because most of outcome-related signals have been captured by IAs or non-activity features, e.g. resource and data.

\begin{figure}
	\centering{\includegraphics[scale=0.7]{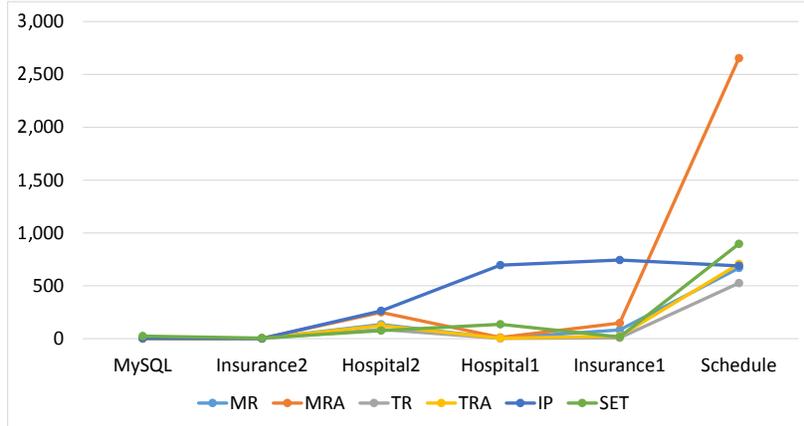}}
	\caption{Mining Runtime (in seconds)}
	\label{fig:runtime}
\end{figure}

\paragraph{Execution times} All experiments were executed on a laptop with 2.5GHz Intel CPU Dual Core, 8GB internal memory, and 64-bit Windows operating system. Fig. \ref{fig:runtime} shows the average feature mining runtime from five-fold iterations. In relation to the dataset characteristics presented in Table \ref{tab:datasets2}, factors affecting mining performance are likely involved with the number of events per case, the number of event classes per case and the data size (total number of cases). This is observed in case of the Schedule dataset with exceptionally long cases, Hospital2 with shortest average cases but largest data size, Insurance1 with quite large data size but low number of event classes per case. Mining techniques respond to different degrees to data characteristics. For example, it can be observed that MRA is highly sensitive to the number of events per case while IP is not; and SET is more susceptible to the number of event classes per case.

\section{Rules Interestingness Evaluation}\label{sec:RulesInterestingnessEval}

We extracted rules from the six datasets across seven feature types using decision tree classifier. This technique takes the path from a leaf node of the tree to the top as a rule. All decision trees are trained with uniform input parameters. As a result, there are seven rulesets for each dataset, one per feature type. The rule has the form of $A=>B$, where the antecedent A is a conjunction of multiple selectors in the form of \textit{attribute OP value} with \textit{OP} as relational operator, and the consequent B is the class label.

For each dataset, we compare the quality of these rulesets on a number of selected measures and observe the effect of pattern types on the rules quality. The measures used in our evaluation include:

\begin{itemize}
	\item Rule Length: the number of operators in the antecedent part of the rule. This is averaged from all rules of a ruleset.
	\item \%Generalization \cite{gallion1993dynamic}: the generalizability based on the number of training examples and the number of rules.
		\begin{equation}\label{eq:generalization}
		\%Generalization = (1 - \frac{\# of Rules}{\# of Training Examples})*100
		\end{equation}
	\item Probability-based Objective Rule Interestingness Measures. \cite{Geng2006} evaluates 38 interestingness measures with reference to 12 desired properties. The authors ranked these measures in terms of how much they hold a property (Table $V$ in \cite{Geng2006}). Based on this evaluation, we were able to select six top measures as listed below (noted that Yule's Q and Yule's Y are normalized variants of Odd Ratio \cite{tan2004select}). Readers should refer to \cite{Geng2006} for a detailed discussion on these measures while a brief description is given below.
	\begin{itemize}
		\item Collective Strength ($\CS$): proposed in \cite{aggarwal98collective} to replace the support-based paradigm in frequent itemset mining. The original idea is to measure the collective strength of an itemset $I$ as $CS(I) = \frac{Good Events}{E[Good Events]}*\frac{E[Bad Events]}{Bad Events]}$, where \textit{Good Events} mean the items of the itemset are present in most of transactions and \textit{Bad Events} are when they are not. In the context of rule evaluation, \textit{Good Events} are when A and B are strongly correlated, and the opposite are \textit{Bad Events}. The collective strength $CS$ ranges from 0 to $+\infty$. If $CS=0$, A and B are mutually exclusive (perfect negative correlation). If $CS=1$, A and B are totally independent. If $CS$ increases to $+\infty$, A and B are increasingly positively correlated. 
		
		\begin{equation}\label{eq:collectivestrength}
		\CS = \frac{P(AB)+P(\neg B|\neg A)}{P(A)P(B)+P(\neg A)P(\neg B)}*\frac{1-P(A)P(B)-P(\neg A)P(\neg B)}{1-P(AB)-P(\neg B|\neg A)}
		\end{equation}
		
		\item Two-Way Support ($\TWS$): proposed in \cite{yao97twoway} to search for interesting knowledge in multiple databases. This measure ranges from $-\infty$ to $+\infty$. If it approaches $+\infty$, the relation $A=>B$ increases and vice versa, the relation $B=>A$ increases when it approaches $-\infty$.
		
		\begin{equation}\label{eq:twowaysupport}
		\TWS = P(AB)*\log_2\frac{P(AB)}{P(A)P(B)}
		\end{equation}
		
		\item $\phico$: this measure is based on Pearson correlation coefficient which ranges between -1 and +1. When it is greater than 0, A and B are positively correlated. When it equals 0, A and B are independent. When it is less than 0, A and B are negatively correlated.
		
		\begin{equation}\label{eq:phicoefficient}
		\phico = \frac{P(AB)-P(A)P(B)}{\sqrt{P(A)P(B)P(\neg A)P(\neg B)}}
		\end{equation} 
		
		\item Piatetsky-Shapiro ($\PS$): originally proposed in \cite{piatetsky91rules}, this measure reflects the variance of probability when A and B are correlated and independent, with values ranging from -0.25 and 0.25. If PS is zero, A and B are totally independent. If PS is greater than 0, A and B are positively correlated, and vice versa, less than 0 when A and B are negatively correlated. 
		
		\begin{equation}\label{eq:piatetsky}
		\PS = P(AB) - P(A)P(B)
		\end{equation}
		
		\item Yule's Q ($\YuleQ$): this is a normalized version of Odds Ratio ($\OR$) \cite{mosteller68odds}, and based on the contingency table involving A and B. Yule's Q ranges from -1 (perfect negative correlation) to +1 (perfect positive correlation).
		
		\begin{equation}\label{eq:oddsratio}
		\OR = \frac{P(AB)P(\neg A\neg B}{P(A\neg B)P(\neg AB)}
		\end{equation}
		
		\begin{equation}\label{eq:yulesQ}
		\YuleQ = \frac{OR-1}{OR+1}
		\end{equation}
		
		\item Information Gain ($\IG$): this measure is related to the Lift measure $\frac{P(AB)}{P(A)P(B)}$, i.e. how many times it needs to lift the confidence from the case when A and B are totally independent $P(A)P(B)$ to the case when A and B are statistically correlated $P(AB)$. IG ranges from $-\infty$ (negatively correlated) to $+\infty$ (positively correlated).
		
		\begin{equation}\label{eq:informationgain}
		\IG = log\frac{P(AB)}{P(A)P(B)}
		\end{equation}
		
	\end{itemize}
\end{itemize}	

\subsection{Patterns and Rules}\label{PatternsAndRules}


Across all datasets and feature types, we observe that patterns with high Fisher score are often chosen to form rules with high coverage based on the number of cases covered. This is justifiable because the Decision Tree would pick up these patterns as preferable splitting points based on their strong discriminative power (Gini index is used for Decision Tree in our experiment). The higher Fisher score a pattern has, the more rules they take part in and the higher coverage those rules are. Several representative examples are reviewed in details as follows.

In particular, IP/Schedule provides a remarkable pattern: "364 198 361" (every number represents one activity). The related high-coverage rule is: \textit{IF "364 198 361" $\geq$ 0.500 THEN failing}, meaning if this pattern occurs, the trace would have a failing outcome. Conversely, no IA-based rules are formed based on these three events, meaning activities individually do not carry any outcome-related signals.

For the Hospital1, a remark is that most rules contain the "Nursing Progress Notes" activity which has the top Fisher score. In this process, "Nursing Progress Notes" seems to dominate the relationship with the duration-based outcome, i.e. if there are more "Nursing Progress Notes", the case would last longer.

For the Insurance1, the IA-based rules basically have similar forms as the IP- and MR-based if composite patterns are broken down into individual activities. Although some IP- and MR-based patterns have slightly higher scores than the best of IAs, for example "1 13 16" (0.344 vs. 0.159) and "1 16" (0.292 vs. 0.159), but it is not substantial as in the case of the Schedule dataset. That's possibly the reason why IP has a better predictive result than other features but cannot surpass IA.

For Insurance2, notably some IP- and SET-based patterns have higher Fisher scores than IA and others. Specifically, there is one remarkable IP-based rule: \textit{IF "Process Additional Information, Contact Customer" $\geq$ 0.5 THEN Rejected Case} (51\% coverage), meaning if "Process Additional Information" and "Contact Customer" occur together, the case would be rejected. We observed that the IA also provides rules relating to "Process Additional Information" and "Contact Customer" but its rules are longer with other activities included. This is likely related to the fact that though "Process Additional Information" and "Contact Customer" rank very high in IA feature list, they are lower than the IP-based patterns.

In this analysis, it is more evident that IAs likely carry most of the outcome-related signals in case of temporal outcome criterion (Hospital1 and Insurance1). Composite patterns, however, contribute some interesting rules (Schedule and Insurance2).

\subsection{Rules Interestingness Measures}

\begin{figure}[htb!]
	\centering
	{\includegraphics[width=\textwidth]{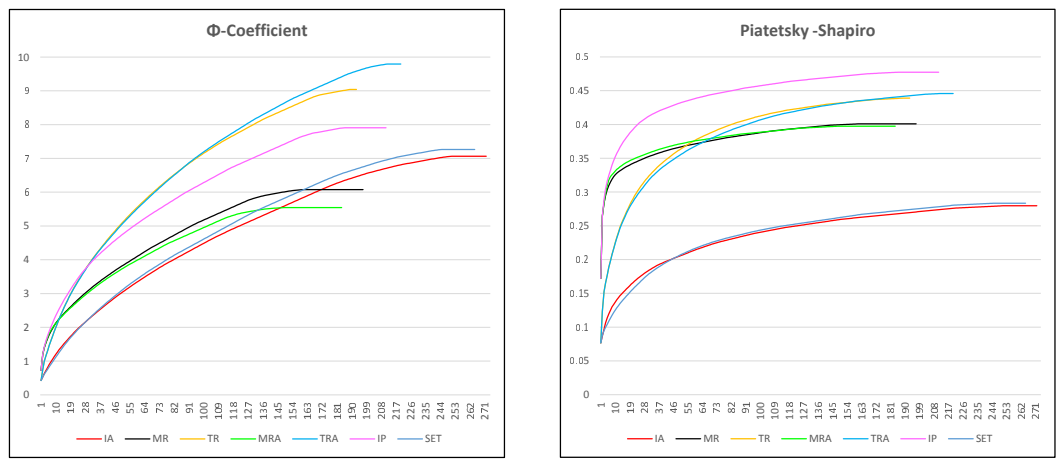}}
	\caption{Schedule dataset - \(\phi\)-Coefficient and Piatetsky-Shapiro}
	\label{fig:interest-schedule}
\end{figure}

\begin{figure}[htb!]
	\centering
	{\includegraphics[width=\textwidth]{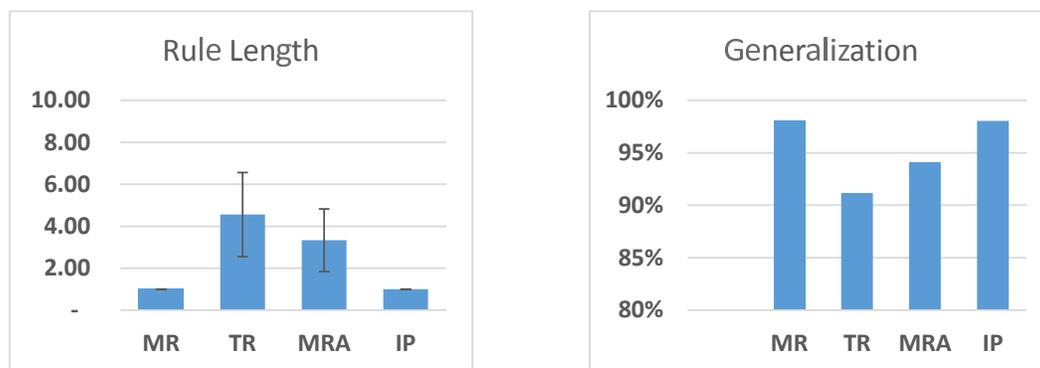}}
	\caption{MySQL dataset - Rule Length \& \%Generalization}
	\label{fig:mysql-length}
\end{figure}

\begin{figure}[!htb]
	\centering
	{\includegraphics[width=\textwidth]{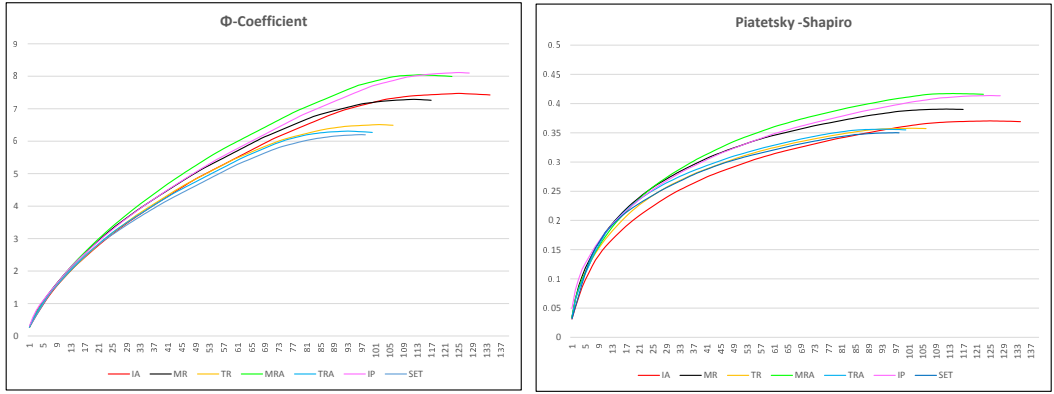}}
	\caption{Hospital1 dataset - \(\phi\)-Coefficient and Piatetsky-Shapiro}
	\label{fig:interest-hospital1}	
\end{figure}

Different metrics are used to evaluate the quality of a ruleset. Rule Length is computed by averaging length of individual rules. \%Generalization is computed according to formula (\ref{eq:generalization}). For probability-based measures, cumulative interestingness is computed by firstly sorting rules within a ruleset in decreasing order of the measure values, then accumulating these values from the highest to lowest. The cumulative measure for one feature type is visualized as an upward curve by the number of rules (Fig. ~\ref{fig:interest-schedule},\ref{fig:mysql-length},\ref{fig:interest-hospital1}).

Our observation is summarized as follows with selected graphs for illustration due to space limitation.

For the Schedule, the IP-based ruleset has consistently high interestingness over the others, except in \(\phi\)-Coefficient where the TR and TRA are the best (Fig.~\ref{fig:interest-schedule}). Notably, IP-based ruleset also achieves the highest predictive outcome, as well as TR and TRA. Rule Length and \%Generalization are quite similar across feature types.

For the MySQL, similar to the Schedule dataset, IP-based and MR-based ruleset show the best cumulative interestingness, which accords with their high predictive strength (Fig.~\ref{fig:mysql-length}). They also have the shortest length and the greatest \%Generalization (Rules were not generated for several feature types due to pruning parameters).

For the Hospital1, the MRA-based and IP-based ruleset have higher cumulative interestingness than others in all measures (Fig.~\ref{fig:interest-hospital1}). They also have high \%Generalization after the IA-based. All rulesets are approximately equal in rule length.

For the Hospital2 dataset, all rulesets are balanced across all measures, which corresponds to their equal predictive power and Fisher score as shown in section \ref{sec:AccuracyEvaluation} and \ref{PatternsAndRules}. For the Insurance1 dataset, similar to the Hospital1 dataset, the MRA-based and IP-based rulesets are higher than others in all interestingness measures. The IP-based and Set-based have slight improvement over others in Rule Length and \%Generalization. For the Insurance2 dataset, the MRA-based cumulatively outperforms others in all measures except \(\phi\)-Coefficient, Yule's Q and Information Gain where the Set-based is the best.

In summary, across all datasets, the IP-based, MRA-based and Set-based ruleset are often at the top of the cumulative interestingness measurement. The IP-based often results in shorter length and higher \%Generalization. The IA-based ruleset is often below the others in most of the measures, except in Yule's Q and Information Gain where it improves after more rules are added to the ruleset. In short, rulesets based on composite patterns has higher interestingness than those based on IA.

\section{Conclusion}\label{sec:Conclusion}
Our review of the state of the art has revealed that existing techniques for business process deviance mining are based on the extraction of patterns from business process execution traces using either individual activity frequency or frequent or discriminative pattern mining approaches. The empirical evaluation has unveiled evidence that, in the context of structured (predictable) business processes, pattern mining approaches clearly outperform the basic approach based on individual activity frequency in terms of accuracy. On the other hand, in the case of business processes with high level of variability, the accuracy improvement obtained by using pattern mining approaches is negligible. The latter observation suggests that when business processes have high variability, no clear composite patterns emerge, and thus such patterns do not add predictive power over simple activity counts. However, we observed that the use of composite patterns (be them frequent or discriminative ones) generally has a positive effect on the interestingness of the mined rule sets.

In the case of processes with high variability, the maximum accuracy (AUC) achieved is around 80\%, which might be insufficient in many practical applications. Underlying this limitation is the fact that the reviewed techniques for business process deviance mining treat the input as consisting of simple symbolic sequences, i.e.\ sequences of simple symbols (tokens) representing in our case activity or event occurrences. In some cases, including all six datasets used in this study, business process execution logs consist instead of \emph{temporal complex} symbolic sequences, i.e.\ sequences of timestamped events, each event with a payload consisting of attribute-value pairs. Such logs can be extracted for example from appropriately instrumented information systems, such as patient management systems (in the case of healthcare processes) or claims handing systems (for insurance claims), or from mainstream Enterprise Resource Planning systems. It is likely that data payloads associated with events can convey significant information regarding possible deviances and thus cannot be ignored as in the techniques evaluated in this paper. 

A direction for future work is thus to develop and apply techniques for extracting (discriminative) patterns from complex symbolic sequences -- a non-trivial and open problem as noted in~\cite{Xing2010Survey}. While tackling the problem of complex symbolic sequence mining in the general case is challenging, it may be possible to reduce the problem of business process deviance mining to well-scoped subsets of this problem, for example by taking advantage of information contained in available process models, such as which activities read/update which data attributes, which data attributes are used to make a decision in the process, etc., in order to prune the pattern search space.

\smallskip\noindent\textbf{Acknowledgments} 
This work is funded by the Estonian Research Council, ERDF via the Estonian Centre of Excellence Programme. This research is also supported by the Australian Centre for Health Services Innovation (\#SG00009-000450) and by the Australian Research Council (\#DP150103356).

\bibliographystyle{ACM-Reference-Format-Journals}
\bibliography{bibliography}





\end{document}